\documentclass[
]{ceurart}

\usepackage{listings}
\usepackage{latexsym}
\usepackage{amssymb}
\usepackage{amsmath}
\usepackage{amsthm}
\usepackage{booktabs}
\usepackage{enumitem}
\usepackage{graphicx}
\usepackage{color}

\usepackage{algorithm}
\usepackage{algorithmic}
\usepackage{subcaption}
\usepackage{pgfplots}
\usepgfplotslibrary{groupplots}
\pgfplotsset{compat=1.18}
\usepackage{tikz}
\usetikzlibrary{patterns,automata, positioning}
\usepackage{xcolor}
\usepackage{mathtools}
\usepackage{hyperref}
\usepackage{cleveref}

\usepackage{paralist}

\newcommand{\Scal}{\mathcal{S}}
\newcommand{\Acal}{\mathcal{A}}
\newcommand{\Tcal}{\mathcal{T}}
\newcommand{\Rcal}{\mathcal{R}}

\newcommand{\seeker}{\textbf{S}}
\newcommand{\helper}{\textbf{H}}

\definecolor{egoblue}{HTML}{4472C4} 
\definecolor{humangreen}{HTML}{70AD47}

\definecolor{humanhumangray}{HTML}{B7B7B7}
\definecolor{agentagentorange}{HTML}{F2A95D}
\definecolor{agentmutebrown}{HTML}{B88A6D}

\definecolor{pptblue}{HTML}{4472C4}

\theoremstyle{definition}

 \newtheorem*{problem*}{Problem}
 
\DeclareMathOperator{\subjectto}{s.t.}

\newcommand{\D}{\mathcal{D}}
\newcommand{\Prop}{\mathrm{Prop}}
\newcommand{\trace}{\sigma}

\newcommand{\C}{\mathcal{C}}

\begin{document}

\copyrightyear{2025}
\copyrightclause{Copyright for this paper by its authors.
  Use permitted under Creative Commons License Attribution 4.0
  International (CC BY 4.0).}

\conference{OVERLAY 2025: 7th Workshop on Artificial Intelligence and Formal Verification, Logic, Automata, and Synthesis}

\title{Learning to Coordinate without Communication under Incomplete Information}

\author[1]{Shenghui Chen}[%
email=shenghui.chen@utexas.edu,
]
\cormark[1]
\address[1]{University of Texas at Austin, USA}

\author[2]{Shufang Zhu}[%
email=shufang.zhu@liverpool.ac.uk,
]
\address[2]{University of Liverpool, UK}

\author[3]{Giuseppe {De Giacomo}}[%
email=giuseppe.degiacomo@cs.ox.ac.uk,
]
\address[3]{University of Oxford, UK}

\author[1]{Ufuk Topcu}[%
email=utopcu@utexas.edu,
]

\cortext[1]{Corresponding author.}

\begin{abstract}
    Achieving seamless coordination in cooperative games is a crucial challenge in artificial intelligence, particularly when players operate under incomplete information. While communication helps, it is not always feasible. 
    In this paper, we explore how effective coordination can be achieved without verbal communication, relying solely on observing each other's actions.
    Our method enables an agent to develop a strategy by interpreting its partner’s action sequences as intent signals, constructing a finite-state transducer built from deterministic finite automata, one for each possible action the agent can take.
    Experiments show that these strategies significantly outperform uncoordinated ones and closely match the performance of coordinating via direct communication.
\end{abstract}


\maketitle

\section{Introduction}\label{sec:intro}
In artificial intelligence, autonomous agents often compete or cooperate, reflecting real-world interactions. Games offer structured settings to study such behaviors.
Much of the research has focused on adversarial games, where agents pursue goals despite adversarial environments~\cite{CimattiRT98,Cimatti03,GeBo2013}. 
Conversely, cooperative games~\cite{dafoe2021cooperative} require agents to collaborate toward a shared goal. In this paper, we are interested in \textit{shared-control games}~\cite{chen2024sharedcontrol}, a form of cooperative games in which two players, the \textit{seeker} and the \textit{helper}, collectively control a single token to achieve a goal. For instance, in robotic warehouses, a human operator~(seeker) navigates to retrieve items while a support robot~(helper) clears obstacles, allowing the operator to progress to its location~(token). Helper agents with such assistive abilities have the potential to enhance collaboration with humans in various settings, from virtual games~\cite{carroll2019utility,bard2020hanabi} to physical applications like assistive wheelchairs~\cite{goil2013using}.

Shared-control games are especially challenging when players have incomplete or differing information. Such asymmetry, from partial observations or limited game understanding, can cause misaligned or suboptimal actions. In robotic warehouses, poor inference can reduce efficiency and pose safety risks.
Direct communication offers a solution by enabling the exchange of relevant information between players. Recent work leverages large language models to express and interpret intentions via natural language, improving coordination in human-AI teams~\cite{guan2023efficient,LiuYG0LWW24,chen2024sharedcontrol}.
However, direct communication is not always feasible due to constraints like limited bandwidth, latency, noise, or task demands. In such cases, coordination must rely on inferring intent from observed behavior alone.

In this paper, we consider scenarios where direct verbal communication is unavailable. In such settings, the helper must infer when assistance is needed based solely on the seeker's trajectory. Our framework generalizes \textit{shared-control games}~\cite{chen2024sharedcontrol} by allowing multi-step control for the seeker and introducing a helper strategy that interprets the observed trajectory for effective coordination.
To obtain a helper strategy, we represent it as a finite-state transducer composed of several deterministic finite automata (DFAs), each corresponding to a specific helper action. Each DFA is learned using a variant of Angluin's L* algorithm~\cite{Angluin87}. The learning process is based on sequences of observed seeker moves, with each DFA accepting those sequences that align with the intention to trigger its associated action and rejecting those that do not. 
The learned DFAs are then combined into a finite-state transducer that encodes the helper’s overall strategy.

We empirically evaluate our proposed solution in \emph{Gnomes at Night}$^{\textit{TM}}$, the same testbed used by \cite{chen2024sharedcontrol}.
We compare the helper's performance in our no-communication coordination approach with two other cases: a worst-case scenario where the helper does not try to coordinate at all, and a best-case scenario where the helper coordinates through direct communication. We measure success rates and the number of steps to complete the game across a given number of trials and different maze configurations. We test on  \(9 \times 9\) and larger \(12 \times 12\) mazes to assess the solution's ability to generalize across maze sizes.
Results show that no-communication coordination with our solution significantly improves success rates over no coordination in both maze sizes and performs comparably to direct communication. It also reduces steps, wall memory, and wall error rate by more than half.
\section{Related Works}

The problem of achieving coordination in multi-agent systems involves enabling autonomous agents to work together toward shared goals. Prior work spans distributed AI \cite{genesereth1988cooperation}, swarm intelligence (stigmergy \cite{marsh2008stigmergic}), and game theory (correlated equilibrium \cite{aumann1974subjectivity}). However, in many settings, all agents are aware of the goal, typically assuming all agents know the goal. In contrast, we study coordination where only one agent knows the goal.

A common approach to address these challenges under incomplete information is through explicit communication, using discrete signals as in Hanabi \cite{bard2020hanabi}, or natural language in negotiation and coordination games like \textit{Deal-or-No-Deal} \cite{lewis-etal-2017-deal,he-etal-2018-decoupling}, \textit{Diplomacy} \cite{paquette2019no}, and \textit{MutualFriends} \cite{he-etal-2017-learning}. Recently, \emph{Gnomes at Night}$^{\textit{TM}}$ was used to highlight the challenges of shared control under incomplete information, when leveraging natural language dialogue for communication \cite{chen2024sharedcontrol}. In contrast, our study examines coordination without direct communication, using a mute version of \emph{Gnomes at Night}$^{\textit{TM}}$.

Another approach to understanding coordination is through multi-agent reinforcement learning (MARL), where agents learn cooperative strategies via trial and error in complex environments, particularly through self-play and opponent modeling \cite{foerster2017learning}. However, most MARL approaches use neural networks to represent policies, which often obscures the intent inference process within the learning model. The automata-learning-based solution technique proposed in this paper provides a more explicit representation, potentially offering better explainability. \textit{Pedestrian trajectory prediction} similarly involves anticipating future actions from past behavior, environmental conditions, and interactions with others---analogous to the helper inferring the seeker’s intent. Approaches include knowledge-based models~\cite{Helbing_1995} and supervised deep learning methods~\cite{AlahiGRRLS16}.

A process mirroring the challenge of the helper attempting to infer the seeker's intended actions is \textit{plan recognition} in planning~\cite{kautz1986generalized}. Goal recognition involves identifying all potential goals an agent might pursue based on a sequence of observed actions~\cite{vilain1990getting,charniak1993bayesian,lesh1995sound,goldman2013new,avrahami2005fast,ramirez2009plan}. In this context, the domain is entirely visible, allowing for the calculation of possible goals that can be achieved through an optimal policy aligning with these observations. However, in our setting, the helper lacks information on the seeker's domain. An efficient coordination could help in the mutual understanding of each player's domain. Exploring how to develop such coordination aligns with the focus of this study.

\section{Preliminaries}

A \emph{deterministic finite automaton (DFA)} is a tuple $\D = (2^\Prop, Q, q_0, \delta, F)$, where $\Prop$ is the alphabet, $Q$ the finite set of states, $q_0 \in Q$ the initial state, $\delta: Q \times 2^\Prop \rightarrow Q$ the transition function, and $F \subseteq Q$ the accepting states. The language $\mathcal{L}(\D)$ denotes the set of accepted traces.

We use \emph{Angluin’s L* algorithm}~\cite{Angluin87} to learn DFAs via two query types: (1) \emph{Membership queries}, where the learner asks whether a trace $\trace$ is accepted; and (2) \emph{Equivalence queries}, where the learner submits a hypothesized DFA and, if incorrect, receives a counterexample to refine it.

\section{Formal Framework}\label{sec:framework} 
We extend the \emph{shared-control game under incomplete information}~\cite{chen2024sharedcontrol} to allow the \textit{seeker} to retain control for multiple steps before transferring it to the \textit{helper}. This modification enables intent to be expressed over action sequences rather than isolated moves.

A shared-control game with seeker multi-step dynamics is defined as a tuple 
$\Gamma = (\Scal, s_{\text{init}}, s_{\text{final}}, \Acal^\seeker, \Acal^\helper, \Tcal^\seeker, \Tcal^\helper)$, 
where $\Scal$ is the finite state space; $s_{\text{init}}$ and $s_{\text{final}}$ are initial and goal states; $\Acal^i$ and $\Tcal^i:\Scal\times \Acal^i\to \Scal$ are the private action sets and deterministic transition functions for each agent $i \in \{\seeker, \helper\}$. 
We extend the seeker’s transition to action sequences via:
\vspace{-0.5em}
\[
\Tcal^\seeker_*(s, [a_1, \dots, a_n]) = \Tcal^\seeker(\dots\Tcal^\seeker(\Tcal^\seeker(s, a_1), a_2), \dots, a_n).
\vspace{-0.5em}
\]
A common reward function $\Rcal: \Scal \times (\Acal^\seeker \cup \Acal^\helper) \to \mathbb{R}$ captures the cooperative objective of minimizing steps to the goal. The seeker $\seeker$ takes the initial turn.

\begin{problem*}\label{problem}
Given $\Gamma$ and a reward function $\Rcal$, the seeker follows a policy $\pi^\seeker: \Scal\times \Acal^\helper\to ({\Acal^\seeker})^+$ unknown to the helper, but whose resulting actions the helper can observe. The goal is to learn a helper policy $\pi^{\helper}:\Scal\times (\Acal^\seeker)^+ \to \Acal^\helper$ that maximizes cumulative reward:
\vspace{-0.5em}
\begin{subequations}
    \begin{flalign}
        \max_{\pi^{\helper}} \quad
        & \sum_{t=0}^T \Rcal(s_t, a_t)\\
        \subjectto \quad
        & \mathbf{a}_0=[], s_0 = s_{\text{init}}, \exists k\in \{0, \ldots, T\} \; s.t. \; s_k = s_{\text{final}}.    \\
        & \begin{cases}
            \mathbf{a}^\seeker_{t+1} = \pi^\seeker(s_t, a^\helper_t) & \text{on $\seeker$'s turn,} \\
            a^\helper_{t+1} = \pi^\helper(s_t, \mathbf{a}^\seeker_t) & \text{on $\helper$'s turn,}
          \end{cases}\quad
          s_{t+1} = \begin{cases}
            \Tcal^\seeker_*(s_t, \mathbf{a}^\seeker_{t+1}) &\text{on $\seeker$'s turn,} \\
            \Tcal^\helper(s_t, a^\helper_{t+1}) &\text{on $\helper$'s turn},
        \end{cases}  
    \end{flalign}
    \vspace{-0.5em}
\end{subequations}
where $t$ indexes turns, and $T$ denotes the total number of turns allowed. 
\end{problem*}
\section{Solution Technique}\label{sec:solution}
The key challenge for the helper is to infer the seeker's required help by observing its action sequences, as direct communication is disallowed. We propose an automata-learning-based approach in which the helper constructs \emph{intent-response DFAs}---one per helper action---to recognize patterns in the seeker’s behavior that imply expected responses. These DFAs, unknown to the seeker, are combined into a finite-state transducer that maps seeker action sequences to helper actions.



\subsection{Learning Helper's Intent-Response DFAs}
The seeker pre-determines a policy $\pi^\seeker$ to express intent through action sequences. The helper must learn to strategically perform the actions expected by the seeker when the seeker cannot proceed. To develop a corresponding strategy $\pi^\helper$ for the helper, we introduce an automata-learning-based technique. The key insight is that when the seeker does not need assistance, it will naturally follow the shortest path. In this case, if the action sequence taken deviates from the shortest path, the extra actions taken are interpreted as intent information. We capture such intent information by associating each helper action with a DFA that accepts such indicative sequences, and use Angluin’s L* algorithm to learn these \emph{intent-response DFAs}. 
The helper plays the role of the learner, querying the seeker (as the teacher) through membership and equivalence queries, learning one DFA per action in parallel.

\smallskip
\noindent\textbf{Membership Query.}
The seeker generates an action sequence, knowing which action it expects the helper to perform. The helper extracts intent segments from the observed sequence, infers an expected action, and performs it.
If the performed action matches the seeker's intent, the seeker replies ``Yes", and all extracted segments are positive examples for the corresponding intent-response DFA $\D^a$~(where $a$ is the helper’s action) and negative for all others. A ``No" indicates negative membership for $\D^a$.

By counterfactual intuition, if no coordination is needed, the seeker would naturally follow the shortest path. Hence, redundancies in the sequence suggest that the seeker's intent is embedded in segments outside this shortest path. To identify these ``informative" segments, the helper constructs a subgraph of visited states, computes the shortest path from prior to current location, and removes it from the action sequence. The remaining segments are hence intent segments.

\smallskip
\noindent\textbf{Equivalence query.} For the equivalence query, it is not feasible for the seeker to compare the learned DFAs with the oracle DFAs it has in mind, as the seeker's strategy $\pi^\seeker$ inherently embeds these oracles. We conduct the equivalence query by querying a bounded number of membership. Once the bound is reached, we conclude that the learned \textit{intent-response DFAs}, denoted as $\textbf{D} = \{\D^{a}\}_{a \in \Acal^\helper}$ where $\D^a = \left\{2^{\Acal^\seeker}, Q^a, q_0^a, \delta^a, F^a\right\}$, for each helper action $a\in\Acal^\helper$, are equivalent to the oracle DFAs.
%
%

\subsection{Helper's Strategy Construction}

The learned intent-response DFAs $\textbf{D}$ allow the helper to recognize the seeker's intent solely by analyzing the seeker's action sequences. When the seeker cannot proceed, it becomes the helper's turn to strategically provide assistance. 
%
%
Given a game $\Gamma = (\Scal, s_{\text{init}}, s_{\text{final}}, \Acal^\seeker, \Acal^\helper, \Tcal^\seeker, \Tcal^\helper)$ and the learned intent-response DFAs $\textbf{D}$, we define a strategy generator, i.e., finite-state transducer $T$, from which we can immediately obtain a helper's strategy $\pi^\helper: \Scal \times (\Acal^\seeker)^+ \rightarrow \Acal^\helper$ to solve the problem in \Cref{problem}, though there is no guarantee of optimality in general.

During the helper’s turn, the helper uses the current state $s$ and the seeker’s previous action sequence $\mathbf{a}^\seeker$ to infer the expected next action. Informative segments are extracted from $\mathbf{a}^\seeker$ and evaluated against each intent-response DFA in $\mathbf{D}$. Accepted actions are filtered by the helper’s transition function, and those with highest frequency are returned as intended actions.
Formally, the strategy generator $T = (\Scal, s_{\text{init}}, \Acal^\seeker, \Acal^\helper, \Tcal^\seeker, \Tcal^\helper, \varrho, \tau)$ is constructed as follows:
\begin{compactitem}
    \item
    $\Scal, s_{\text{init}}, \Acal^\seeker, \Acal^\helper, \Tcal^\seeker, \Tcal^\helper$ are the same as in $\Gamma$.
    \item
    $\varrho \colon \Scal \times \mathbf{a}^\seeker_{t+1} \rightarrow 2^\Scal$ is the transition function, where \(\mathbf{a}^\seeker_{t+1} = [a^\seeker_{t_1}, \dots, a^\seeker_{t_n}]\) is the observed seeker's action sequence, such that $\varrho(s, \mathbf{a}^\seeker_{t+1}) = \{\Tcal^\helper (s, a^\helper) \mid a^\helper \in \tau(s, \mathbf{a}^\seeker_{t+1})\}$.
	\item 
	$\tau \colon \Scal \times \mathbf{a}^\seeker_{t+1} \rightarrow 2^{\Acal^{\helper}}$ is the output function such that $\tau(s, \mathbf{a}^\seeker_{t+1}) = \text{NCC}(s, \mathbf{a}^\seeker_{t+1}, \Acal^\helper, \Tcal^\helper, \textbf{D})$. See \Cref{alg:no_comm_coord}.
\end{compactitem}

\begin{algorithm}[t]
\caption{No-Communication Coordination (NCC)}
\label{alg:no_comm_coord}
\textbf{Input}: current state $s$, seeker action sequence $\mathbf{a}^\seeker$, action space $\Acal^\helper$, transition function $\Tcal^\helper$,  intent-response DFAs $\textbf{D}$\\
\textbf{Output}: a set of helper actions $A^\helper$
\begin{algorithmic}[1] 
    \STATE Initialize frequency count $f(a) = 0$ for all $a \in \Acal^\helper$
    \STATE $\{\tau_1, \tau_2, \dots\} \gets \texttt{Capping}(\mathbf{a}^\seeker)$ 
    \FOR{each segment $\tau_i$ and each action $a \in \Acal^\helper$}
        \IF{$\D^a$ accepts $\tau_i$}
            \STATE $f(a) \gets f(a) + 1$
        \ENDIF
    \ENDFOR
    \STATE Set $f(a)=0$ where $\Tcal^\helper(s,a)$ is invalid
    \STATE \textbf{return} a set of actions with the maximum frequency $A^\helper = \arg\max_{a\in \Acal^\helper} f(a)$
\end{algorithmic}
\end{algorithm}



This construction avoids the exponential blowup of DFA composition by evaluating each DFA independently on extracted segments. Hence, the transducer size is linear in the size of the DFAs, and the cost of obtaining intended actions is also linear in $|\Acal^\helper|$.
$T$ generates a strategy by allowing the helper to arbitrarily select an action returned by the output function $\tau(s, \mathbf{a}^\seeker)$, which provides all equally likely intended actions. The strategy is non-Markovian, as $\tau$ depends on the full seeker sequence rather than just the last state or action.

It is worth noting that every helper's intent-response DFA $\D^a$ in $\textbf{D}$ is defined only based on the seeker's actions. Consequently, as long as the seeker utilizes the same policy $\pi^\seeker$ to express its intentions, we can apply these DFAs $\textbf{D}$ across various games that share the same action space of both players.

\section{Simulation Experiment} 

\begin{figure}[h]
    \centering
    \includegraphics[trim={2cm 0 2cm 0},clip,width=0.6\linewidth]{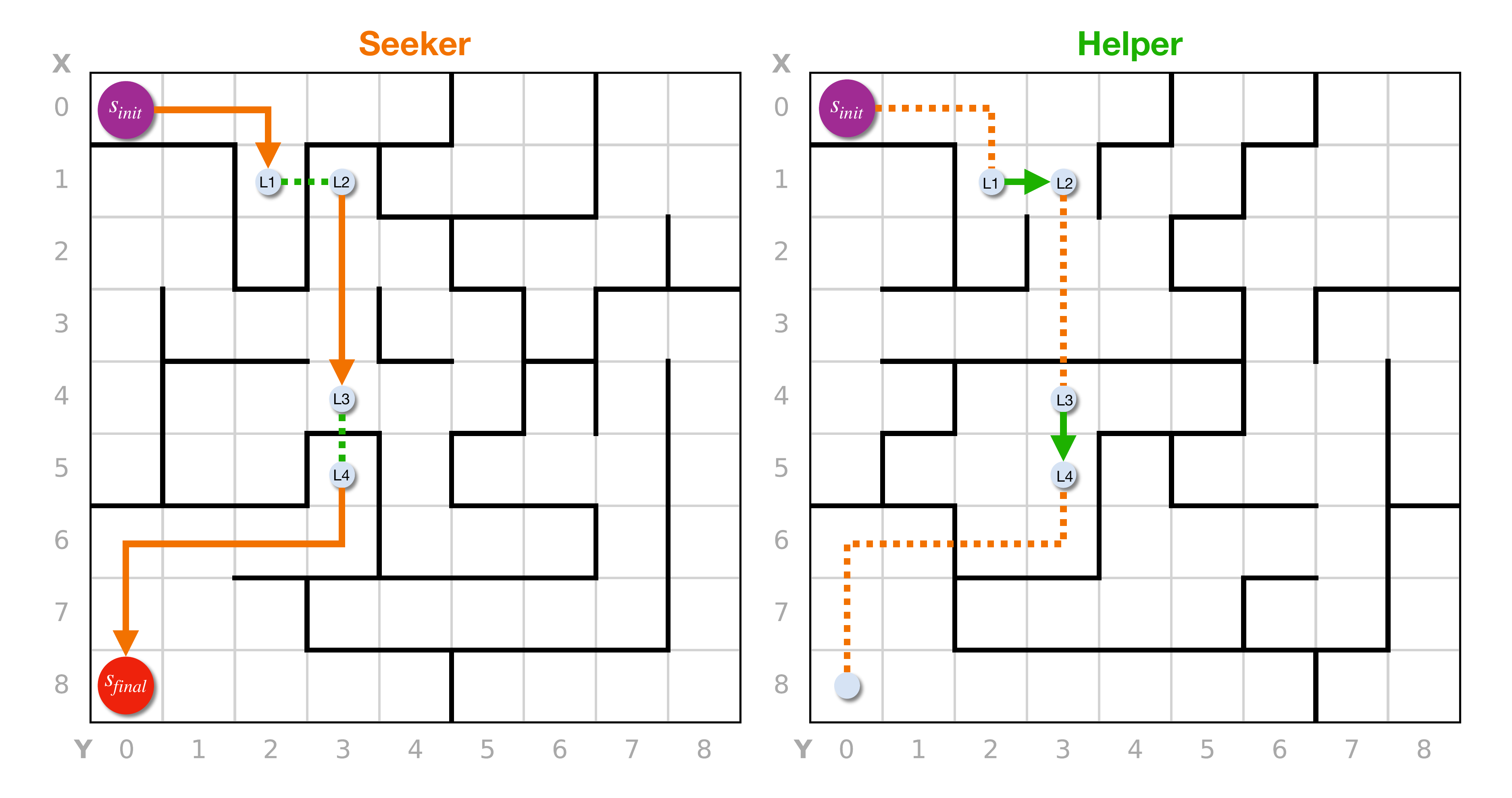}
    \caption{Shared-control game in \emph{Gnomes at Night}$^{\textit{TM}}$: Purple dot marks token at $s_{\text{init}}$, red dot marks $s_{\text{final}}$, and maze layouts captured in $\Tcal^\seeker$ (left) and $\Tcal^\helper$ (right).}
    \label{fig:game}
\end{figure}

\smallskip\noindent\textbf{Gnomest at Night Testbed.}
We illustrate the coordination challenge in shared-control games with incomplete information using the setup shown in \Cref{fig:game}. The left board displays the feasible moves for the seeker, while the right board shows those for the helper. Notably, each player is constrained by their own set of walls, leading to distinct feasible moves for each. In the example shown in~\Cref{fig:game}, the token starts in the top-left corner and must reach the bottom-left goal state. However, the seeker begins inside a T-shaped enclosure that prevents independent progress, making cooperation with the helper essential.
For example, when the token is at L1, the helper must move right to L2 to free the seeker from the enclosure. Later, at L3, the helper must move down to L4 so the seeker can continue toward the goal. 

\smallskip\noindent\textbf{Configurations.}
We evaluate our approach in the \emph{Gnomes at Night}\textsuperscript{TM} testbed~\cite{chen2024sharedcontrol}, where each configuration consists of a maze layout and a treasure location~(see \Cref{appendix:example}).
To assess generalization, we use 10 unseen layouts each for \(9 \times 9\) and \(12 \times 12\) mazes, each with 5 distinct treasure positions, 50 configurations per size. Experiments were run on a MacBook Pro (Apple M1, 8GB RAM, Python 3.9+).

\smallskip\noindent\textbf{Baselines.}
We evaluate three coordination types with three different levels of information exchange: 
In the \textbf{no coordination~(NC)} setting, the seeker plans its path using a modified A* algorithm (see \Cref{algo:modified_a_star} in \Cref{appendix:seeker_implement}), while the helper attempts to guess the seeker's desired next action, but due to a lack of communication, its actions are essentially random. 
With \textbf{direct communication coordination~(DCC)}, both players have a clear communication channel, allowing the seeker to directly inform the helper of its desired help, which the helper then executes on its turn. 
In our proposed \textbf{no communication coordination~(NCC)} setting, the seeker incorporates its required help into its trajectory using the proposed DFA-based approach. The helper interprets the seeker's trajectory and chooses its next action based on the perceived intent. 
For all conditions, the seeker implementation remains, and only the helper strategy varies.

Each coordination type is evaluated with \(n = 100\) trials per configuration. A trial is considered successful if either agent reaches the treasure within \(m = 300\) steps (for \(9 \times 9\)) or \(m = 600\) steps (for \(12 \times 12\)); otherwise, it is marked as a failure.


\smallskip\noindent\textbf{Seeker and Helper Implementations}
The seeker plans paths with a modified A* algorithm that minimizes wall violations under its own and inferred partner constraints. Upon violation, it replans and inserts intent-expressive actions. Deviations by the helper trigger belief updates about unknown walls. See \Cref{appendix:seeker_implement,appendix:helper} for details.

To train the helper, we collect 100 trajectories from $9\times9$ mazes and use the L* algorithm to learn intent-response DFAs for \texttt{right}, \texttt{up}, \texttt{left}, and \texttt{down}. Average learning time is under 0.4s. Jaccard similarity \cite{jaccard1901etude} with oracle DFAs ranges from 0.58 to 0.80. 

\begin{figure*}[h]
    \centering
    \begin{tikzpicture}
    \begin{groupplot}[
        group style={
            group size=2 by 1,
            horizontal sep=2cm,
        },
        height=6cm,
        width=0.4\linewidth,
        ylabel near ticks,
    ]
        \nextgroupplot[
            ybar,
            bar width=0.6cm,
            ylabel={Success Rate (\%)},
            ymajorgrids = true,
            scaled y ticks = false,
            ymin=0,
            ymax=110,
            ytick={0,20,40,60,80,100},
            symbolic x coords={$9\times9$ Maze,$12\times12$ Maze},
            xtick=data,
            enlarge x limits=0.5,
            major x tick style = transparent,
            nodes near coords,
            every node near coord/.append style={font=\scriptsize, anchor=south, yshift=2pt},
            legend style={
                at={(1.2,1.1)},
                anchor=south,
                legend columns=-1, 
                column sep=1ex
            }
        ]
            \addplot+[] coordinates {
                ($9\times9$ Maze, 28.62)
                ($12\times12$ Maze, 17.70)
            }; 
            \addlegendentry{No Coordination (NC)};
                         
            \addplot+[] coordinates {
                ($9\times9$ Maze, 90.16)
                ($12\times12$ Maze, 90.54)
            }; 
            \addlegendentry{No-Communication Coordination (NCC)};
    
            \addplot+[] coordinates {
                ($9\times9$ Maze, 94.08)
                ($12\times12$ Maze, 98.02)
            }; 
            \addlegendentry{Direct-Communication Coordination (DCC)};
                        
        \nextgroupplot[
            ybar,
            bar width=0.6cm,
            ylabel={Steps Taken},
            ymajorgrids = true,
            scaled y ticks = false,
            ymin=0,
            ytick={0,100,200,300,400,500,600},
            symbolic x coords={$9\times9$ Maze,$12\times12$ Maze},
            xtick=data,
            enlarge x limits=0.5,
            major x tick style = transparent,
        ]
            \addplot+[
                error bars/.cd,
                y dir=both,
                y explicit
            ] coordinates {
                ($9\times9$ Maze, 225.94) +- (0, 36.28) 
                ($12\times12$ Maze, 506.49) +- (0, 92.21) 
            };
    
            \addplot+[
                error bars/.cd,
                y dir=both,
                y explicit,
            ] coordinates {
                ($9\times9$ Maze, 69.80)  +- (0, 10.08)
                ($12\times12$ Maze, 102.83) +- (0, 17.90)
            };
    
            \addplot+[
                error bars/.cd,
                y dir=both,
                y explicit
            ] coordinates {
                ($9\times9$ Maze, 36.66)  +- (0, 3.54)
                ($12\times12$ Maze, 34.26) +- (0, 3.36)
            };
    \end{groupplot}
    \end{tikzpicture}  
    \caption{Success rates (left) and steps taken (right) compared across 50 different configurations for $9\times 9$ and $12\times 12$ mazes, illustrating performance across No Coordination (NC), No-Communication Coordination (NCC), and Direct-Communication Coordination (DCC).}
    \label{fig:metrics_bar_plot}
    \vspace{-1em}
\end{figure*}

\smallskip\noindent\textbf{Metrics.}
We report three metrics for each coordination type: (1) \textbf{Success rate}, defined as the fraction of successful trials averaged over 50 configurations; (2) \textbf{Steps taken}, reported as the mean and standard deviation of steps to termination across trials and configurations; and (3) \textbf{Seeker memory}, evaluated by comparing the seeker’s memorized wall constraints with the helper’s actual maze layout, reporting the mean and standard deviation of both the number of memorized walls and their error rate.

\smallskip\noindent\textbf{Hypotheses.}
\textbf{(H1)} NCC outperforms NC in success rate, but underperforms DCC.  
\textbf{(H2)} NCC yields fewer steps than NC, but more than DCC.  
\textbf{(H3)} NCC lowers both the number and error rate of memorized walls compared to NC.
\vspace{-0.5em}
\subsection{Results}
\vspace{-0.5em}
\begin{table*}
    \centering
    \caption{Number of walls memorized by the seeker and the error rate in its memory across 50 configurations per maze size, comparing No Coordination (NC) and No-Communication Coordination (NCC).}
    \begin{tabular}{lr@{\hspace{1mm}}c@{\hspace{1mm}}rr@{\hspace{1mm}}c@{\hspace{1mm}}rr@{\hspace{1mm}}c@{\hspace{1mm}}rr@{\hspace{1mm}}c@{\hspace{1mm}}r}
        \toprule
         & \multicolumn{6}{c}{$9\times 9$ Maze} & \multicolumn{6}{c}{$12\times 12$ Maze}  \\
         \cmidrule(r){2-7} \cmidrule(r){8-13}
        \textbf{Memorized Metric}  & \multicolumn{3}{c}{NC} & \multicolumn{3}{c}{\textbf{NCC}} & \multicolumn{3}{c}{NC} & \multicolumn{3}{c}{\textbf{NCC}} \\
        \midrule
        Number of Walls     & $23.27$ & $\pm$ & $2.23$  & $\mathbf{11.79}$ & $\mathbf{\pm}$ & $\mathbf{1.48}$ & $40.35$ & $\pm$ & $3.64$ & $\mathbf{12.18}$ & $\mathbf{\pm}$ & $\mathbf{1.56}$ \\
        Wall Error Rate (\%)  & $32.38$ & $\pm$ & $12.31$  & $\mathbf{14.17}$ & $\mathbf{\pm}$ & $\mathbf{12.36}$ & $44.39$ & $\pm$ & $11.02$ & $\mathbf{17.42}$ & $\mathbf{\pm}$ & $\mathbf{21.55}$\\
        \bottomrule
    \end{tabular}
    \label{tab:wall_memory}
    \vspace{-1em}
\end{table*}


\smallskip\noindent\textbf{On H1 (Success Rate).}
The left plot in \Cref{fig:metrics_bar_plot} shows that NCC significantly outperforms NC, improving success rates by $61.54\%$ (\(9 \times 9\)) and $72.84\%$ (\(12 \times 12\)). NCC approaches oracle-level performance, with success rates within $4$-$7$\% of DCC. 
A Mann-Whitney U test \cite{mann1947test} confirms NCC significantly outperforms NC (\(p < 0.001\)) in both sizes, while differences between NCC and DCC are not statistically significant~(\(p > 0.1\)). 
These results not only support H1 but surpass our initial expectations.

\smallskip\noindent\textbf{On H2 (Steps Taken).}
The right plot in \Cref{fig:metrics_bar_plot} shows that NCC reduces steps compared to NC in both maze sizes (\(p < 0.001\)), but requires more steps than DCC (\(p < 0.001\)), as expected as NCC requires more steps to effectively express its intentions through its trajectory
These results support H2.

\smallskip\noindent\textbf{On H3 (Seeker Memory).}
\Cref{tab:wall_memory} shows NCC reduces both constraint count and error rate versus NC: by $49.4\%$ and $56.2\%$ in \(9 \times 9\), and $69.8\%$ and $60.8\%$ in \(12 \times 12\), respectively. 
These results support H3, showing NCC minimizes unnecessary exploration and improves intent identification efficiency. 



\section{Conclusion and Future Work}
We studied how a helper agent can learn to coordinate with a seeker in cooperative games without communication. Our approach uses automata learning to infer the seeker's intent by constructing a DFA for each helper action. Experiments in \emph{Gnomes at Night}\textsuperscript{TM} show that this method approaches the performance of an oracle with direct communication.

Future work includes developing an iterative version that refines the helper’s strategy over time, extending from standard reachability to temporal objectives, and adapting to settings with greater non-determinism, such as human or environmental interactions.






\section*{Declaration on Generative AI}
 During the preparation of this work, the author(s) used GPT-4o for: Grammar and spelling check, paraphrasing and rewording, and improving writing style. After using this tool, the authors reviewed and edited the content as needed and take full responsibility for the publication’s content.

\bibliography{ref}

\newpage
\appendix
\textit{All references cited in this appendix are included in the main reference list above.}
\smallskip

\section{Seeker Implementation}\label{appendix:seeker_implement}

The seeker implementation is based on the A* algorithm to generate a path from its current position to the goal. A* algorithm is a widely used graph traversal algorithm that finds the shortest path in a graph by combining the actual distance from a start node to a goal. It selects the path that minimizes \( f(n) = g(n) + h(n) \), where \( g(n) \) is the cost from the start node to \( n \), and \( h(n) \) is the heuristic estimate from \( n \) to the goal. A* guarantees optimal and complete solutions as long as the heuristic does not overestimate the actual cost \cite{hart1968formal}. 


\begin{algorithm}[ht]
\caption{Minimum Violation A* with Hard Constraints}
\label{algo:modified_a_star}
\textbf{Input}: current state $s_0$, final state $s_{\text{final}}$, seeker actions $\Acal^\seeker$, seeker transition function $\Tcal^\seeker$\\
\textbf{Parameter}: wall constraints $\C$, violation cost $c_v$ \\
\textbf{Output}: seeker action sequence $\mathbf{a}^\seeker$
\begin{algorithmic}[1] 
\STATE Initialize priority queue $H$ with $(0, s_0, 0, 0, [])$
\STATE Set \texttt{cost}$[s_0] \gets 0$ and \texttt{violations}$[s_0] \gets 0$
\WHILE{$H$ is not empty}
    \STATE Pop $(\_, s, c, v, \mathbf{a})$ from $H$
    \IF{$s = s_{\text{final}}$}
        \STATE \textbf{return} $\mathbf{a}$
    \ENDIF
    \FOR{each $a \in \Acal^\seeker$}
        \STATE $s' \gets$ \texttt{next}$(s, a)$
        \IF{$s'$ is in-bound and does not violate $\C$}
            \STATE $c' \gets c + 1$
            \IF{$s' \notin [\Tcal^\seeker(s,a) \forall a\in\Acal^\seeker]$}
                \STATE $v' \gets v + 1$
            \ELSE
                \STATE $v'\gets v$
            \ENDIF
            \IF{$s' \notin \texttt{cost}$ \textbf{or} $c' < \texttt{cost}[s']$ \textbf{or} ($s' \in \texttt{violations}$ \textbf{and} $v'<\texttt{violations}[s']$)}
                \STATE \texttt{cost}$[s'] \gets c'$, \texttt{violations}$[s'] \gets v'$
                \STATE $f \gets c' + v' \cdot c_v + \texttt{manhattan}(s', s_f)$
                \STATE Push $(f, s', c', v', \mathbf{a} + [a])$ to $H$
            \ENDIF
        \ENDIF
    \ENDFOR
\ENDWHILE
\STATE \textbf{return} $[]$ 
\end{algorithmic}
\end{algorithm}

\begin{wrapfigure}{r}{0.35\linewidth}
    \centering
    \captionsetup{skip=0pt,format=plain}
    \includegraphics[trim={0 0 30cm 0},clip,width=\linewidth]{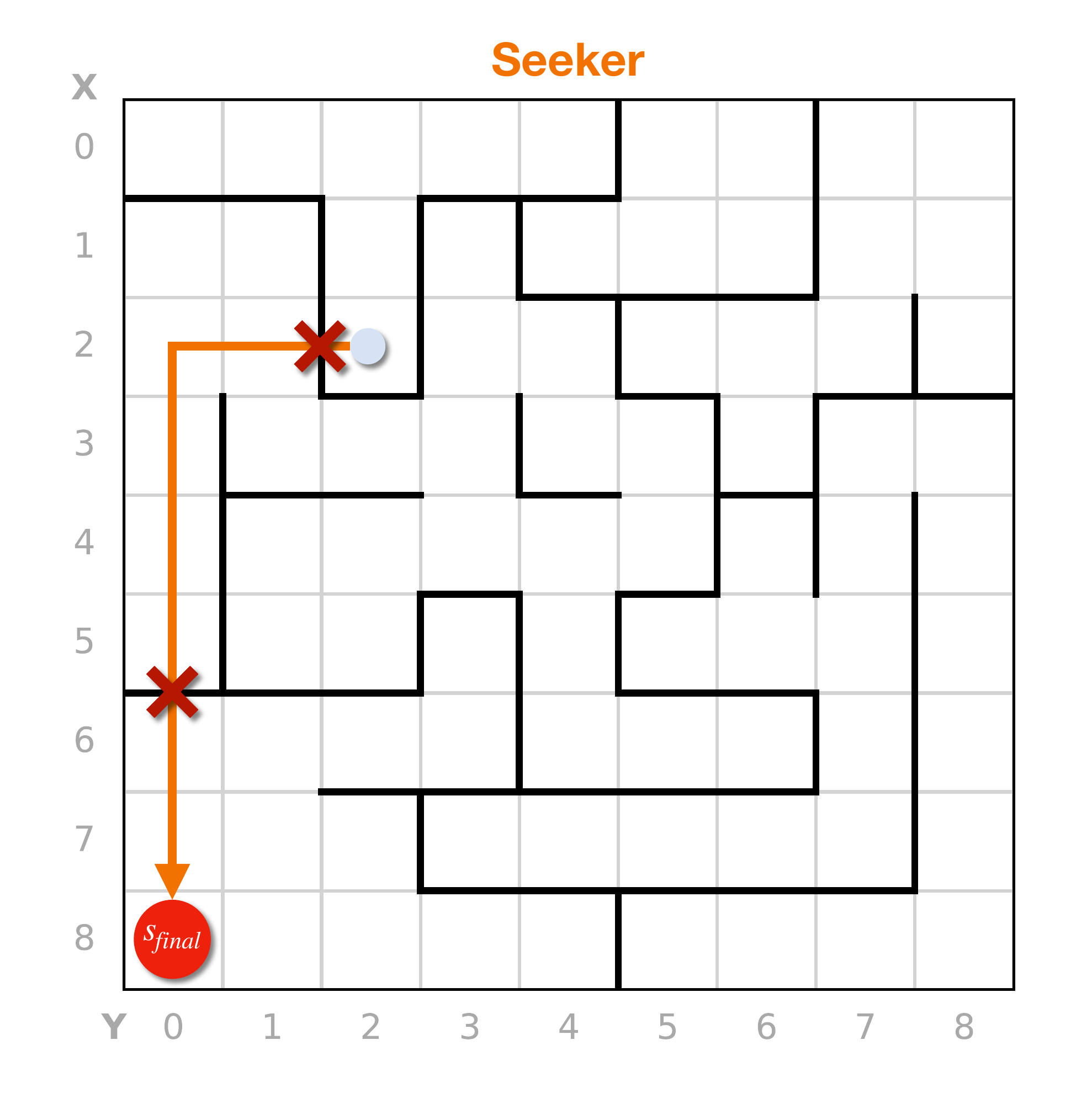}
    \caption{Seeker behavior in the maze: the orange line shows the path from \Cref{algo:modified_a_star}; red crosses mark wall crossings.}
    \label{fig:shortest_path}
\end{wrapfigure}

In particular, the seeker utilizes a modified A* algorithm (\Cref{algo:modified_a_star}) to generate a path that leads to the goal while minimizing wall violations and considering walls that it believes are present for its partner as hard constraints. These constraints are strictly enforced during replanning. We show an example of such a generated path in \Cref{fig:shortest_path}. The algorithm produces an action sequence marked with violation points, and the seeker follows the sequence until the next wall violation. The seeker then inserts intent-expressing segments to indicate the next action it desires from the helper to cross the wall. If the seeker notices the helper's actions do not match its intended actions, it assumes an unknown wall is blocking the helper and adds this wall to its memory as a new hard constraint.

\Cref{algo:modified_a_star} operates similarly to the standard A* algorithm, using a priority queue $H$ to track which paths to explore next. The algorithm starts by adding the current state $s_0$ to the queue with an initial priority of $0$ (line 1). Two dictionaries, \texttt{cost} and \texttt{violations}, are initialized to store the minimum cost and violations needed to reach each state from the start (line 2).
The algorithm then enters a loop, selecting the path with the lowest combined cost and violations from $H$ (line 4). If the current state is the final state $s_{\text{final}}$, the algorithm returns the action sequence taken (lines 5-6).
For each possible action, the algorithm checks if the move stays within bounds and does not violate any hard constraints $\C$ stored in memory. For valid actions, it calculates the new path cost $c'$. If this new path to state $s'$ is either cheaper in cost or with fewer violations than previously known paths to that state, the algorithm updates \texttt{cost} and \texttt{violations}. It then calculates a new priority $f$ for the path, which includes the new cost $c'$, the number of violations $v'$ multiplied by a violation cost $c_v$, and the Manhattan distance from the current state to the final state (line 19). The updated action sequence, along with its priority, is added to $H$ for future exploration (line 20).
This process continues until the algorithm finds a valid path to the final state or exhausts all possibilities. If no valid path exists under the given constraints, the algorithm returns an empty list, indicating that the goal is unreachable.


\section{Helper Implementation}\label{appendix:helper}
We recorded $100$ trajectories (10 per maze) from ten distinct \(9\times9\) layouts.  
From these trajectories, we identify the ``informative" segments and apply the approach from \Cref{sec:solution} to learn intent-response DFAs for the actions \{\texttt{right}, \texttt{up}, \texttt{left}, \texttt{down}\} using the L* algorithm in the Python \href{https://github.com/steynvl/inferrer}{\textit{inferrer}} library.  
Averaged over ten runs per action, learning completed in
\(0.358\!\pm\!0.061\) s (\texttt{right}),
\(0.084\!\pm\!0.018\) s (\texttt{up}),
\(0.167\!\pm\!0.026\) s (\texttt{left}),
and \(0.145\!\pm\!0.034\) s (\texttt{down}).

We also compare the intent-response DFAs against the oracle DFAs the seeker has in mind by calculating their Jaccard similarity \cite{jaccard1901etude}, a statistical measure that assesses the similarity between two sets by dividing the size of their intersection by the size of their union. We enumerate the strings of both DFAs to the same length as two sets to calculate this index. The results indicate Jaccard similarities of $0.8$, $0.58$, $0.67$, and $0.58$ for the DFAs corresponding to the actions \texttt{right}, \texttt{up}, \texttt{left}, and \texttt{down}, respectively. 
Visualizations of the learned DFAs can be found in \Cref{appendix:learned_dfa}.


Although the transducer is trained solely on \(9\times9\) data (as in \cite{chen2024sharedcontrol}), it generalizes to larger mazes. Since L* learns directly from observed behavior, the helper remains effective even when the seeker follows sub-optimal heuristics (e.g., weighted A*), inferring DFAs that capture actual, potentially noisy, behavioral patterns.

\section{How closely the inferred intent DFAs reflect the pre-defined coordination mechanism?}\label{appendix:learned_dfa}


We can express the pre-defined coordination mechanisms as DFAs. 
We compared the learned DFAs with these ground truth DFAs using the Jaccard similarity index, a statistical measure that compares the similarity between two sets by calculating the size of their intersection divided by the size of their union.

\begin{figure*}[ht]
    \centering
    \begin{minipage}[b]{0.45\textwidth}
        \centering
        \includegraphics[width=\textwidth]{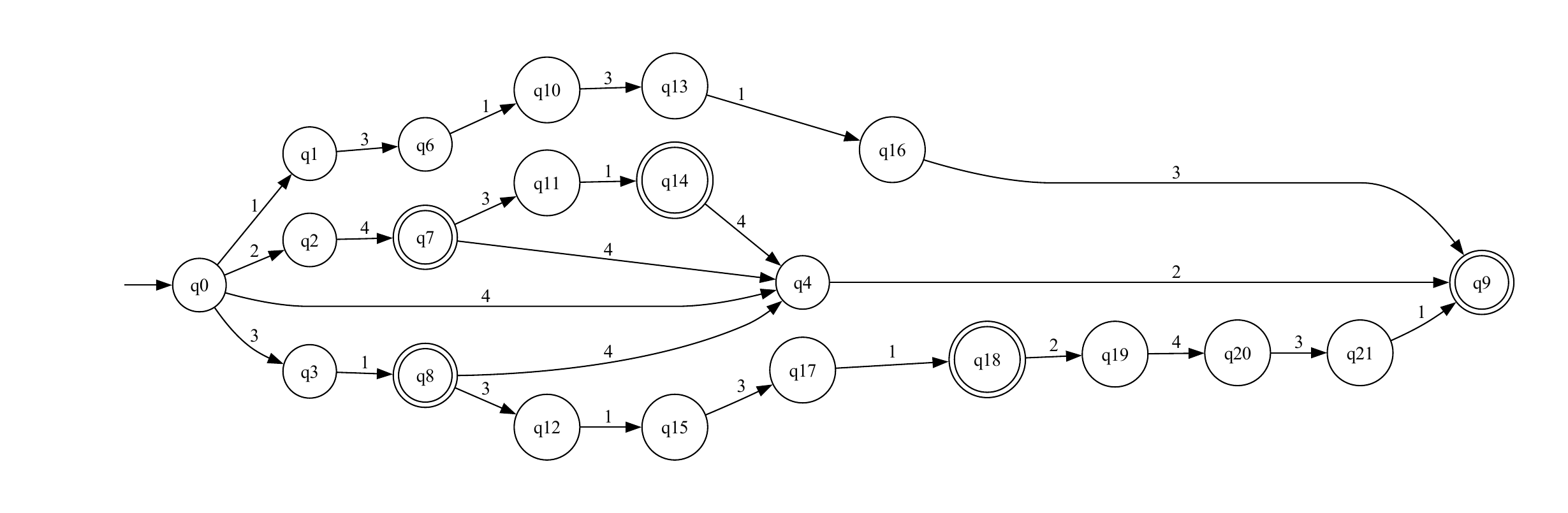}
        \caption{DFA inferred for action \texttt{right} \\(Jaccard similarity: $0.80$)}
    \end{minipage}
    \begin{minipage}[b]{0.45\textwidth}
        \centering
        \includegraphics[width=\textwidth]{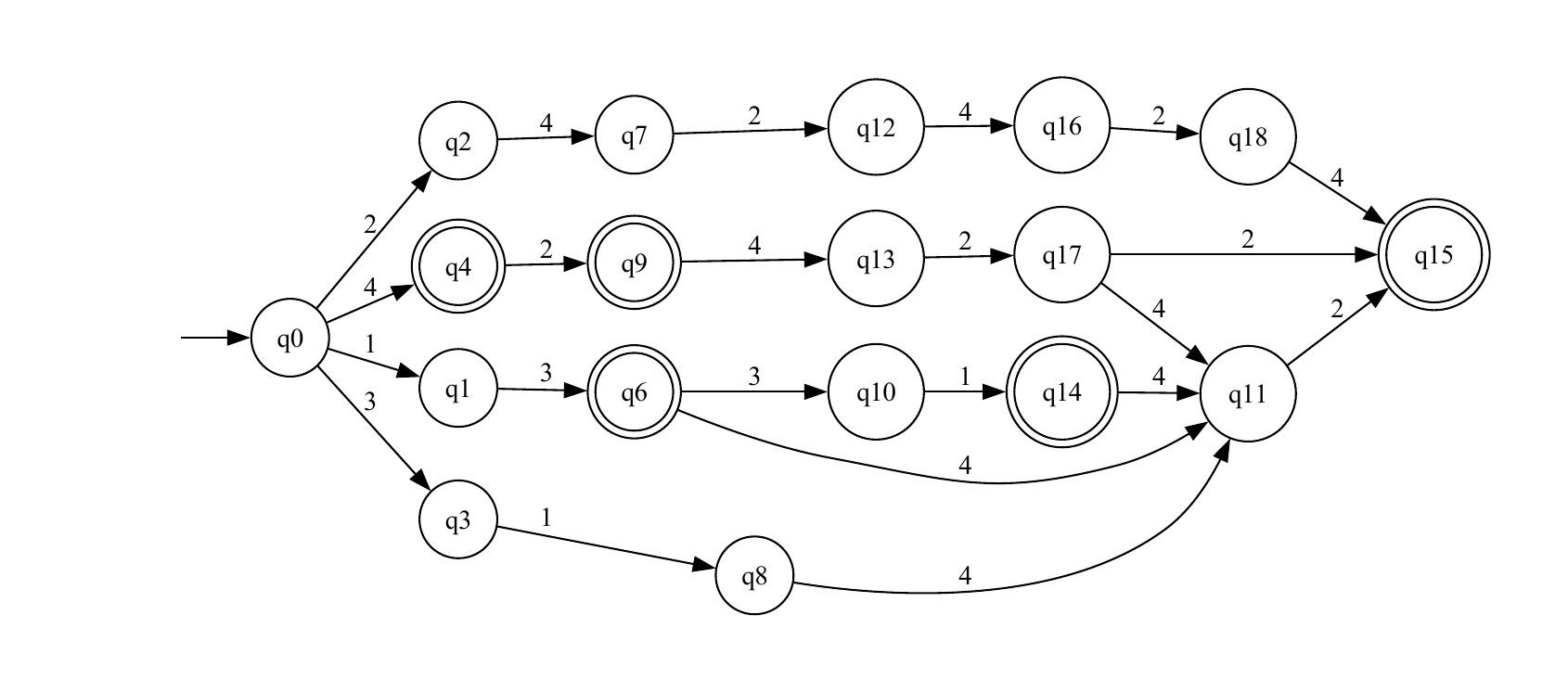}
        \caption{DFA inferred for action \texttt{up} \\(Jaccard similarity: $0.58$) }
    \end{minipage}
    
    \vspace{0.5cm} 

    \begin{minipage}[b]{0.45\textwidth}
        \centering
        \includegraphics[width=\textwidth]{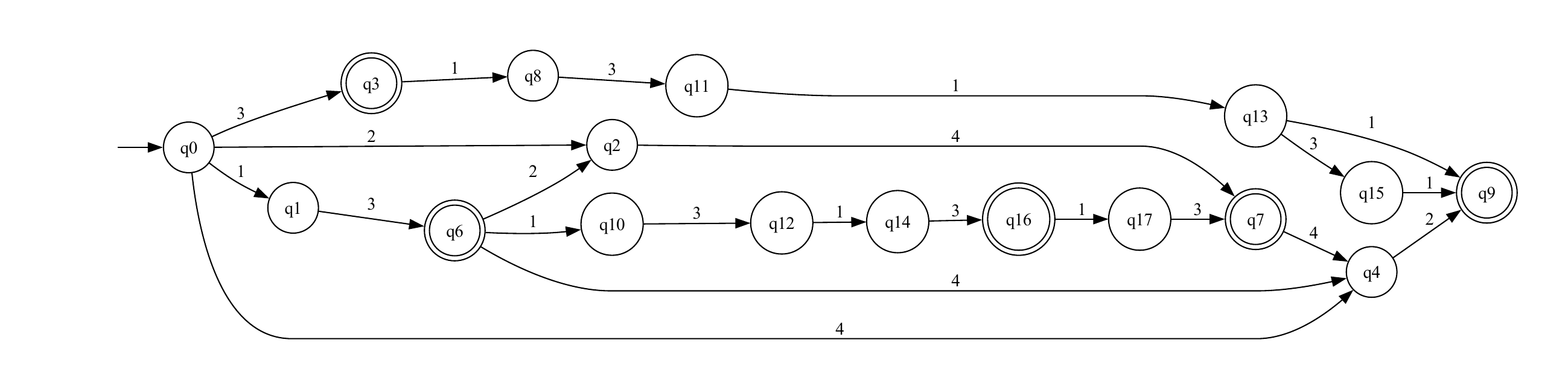}
        \caption{DFA inferred for action \texttt{left} \\(Jaccard similarity: $0.67$)}
    \end{minipage}
    \begin{minipage}[b]{0.45\textwidth}
        \centering
        \includegraphics[width=\textwidth]{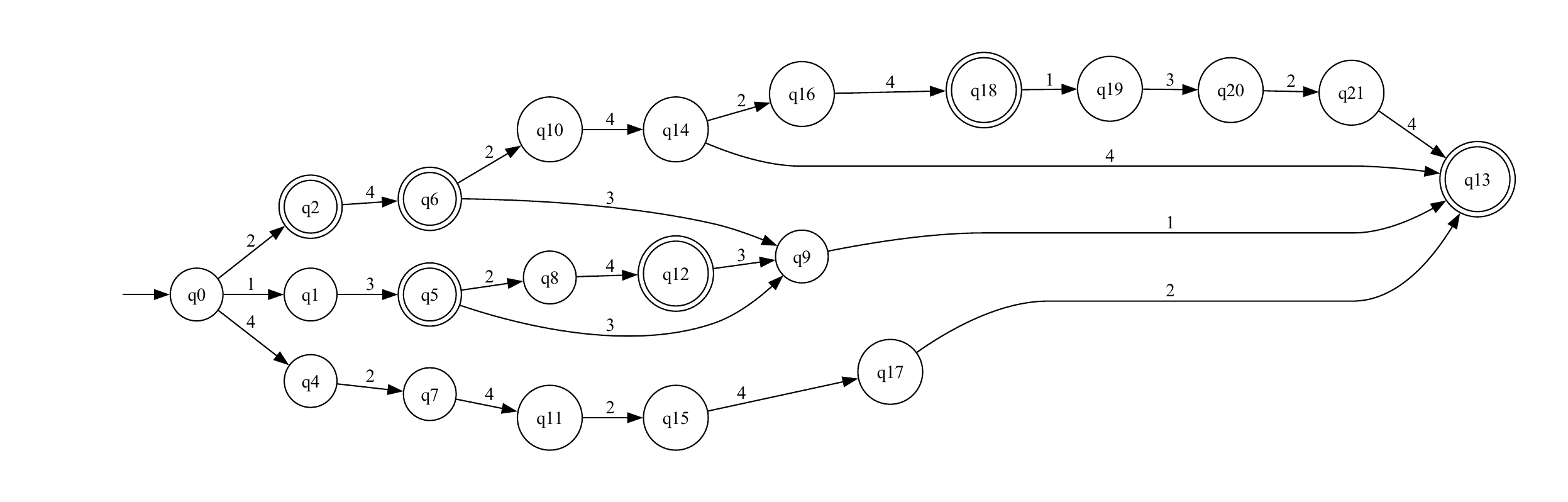}
        \caption{DFA inferred for action \texttt{down} \\(Jaccard similarity: $0.58$)}
    \end{minipage}
    \caption{DFA visualization with Jaccard Similarities}
    \label{fig:dfa_jaccard}
\end{figure*}

\Cref{fig:dfa_jaccard} presents the inferred DFAs for the actions: right, up, left, and down, with their Jaccard similarity scores in captions. The \texttt{right} action shows the highest similarity of $0.8$, suggesting strong alignment with the intended strategy, while \texttt{up} and \texttt{down} display the lowest scores of $0.58$ among the four actions. These results highlight which actions are better represented by the inferred DFAs and identify where additional data sampling is needed. 


\end{document}